Archil Avaliani

International University

December 1, 2002

# *Quantum Computers*

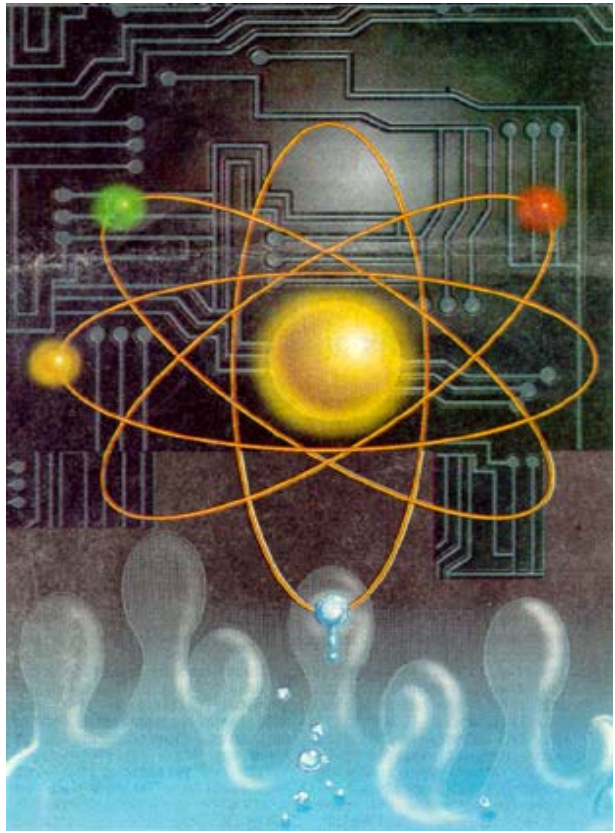

**Picture From USA Today**

On April 18, 1998 Neil Gershenfeld (MIT) and Isaac Chuang (IBM Almaden Research Center) announced about the construction of the first quantum computer. Regardless to its small size it could successfully demonstrate a quantum search algorithm.

## Abstract

This research paper gives an overview of quantum computers – description of their operation, differences between quantum and silicon computers, major construction problems of a quantum computer and many other basic aspects. No special scientific knowledge is necessary for the reader.

## Introduction to Quantum Computers

Around 2030 computers might not have any transistors and chips. Think of a computer that is much faster than a common classical silicon computer. This might be a quantum computer. Theoretically it can run without energy consumption and billion times faster than today's PIII computers. Scientists already think about a quantum computer, as a next generation of classical computers.

Gershenfeld says that if making transistors smaller and smaller is continued with the same rate as in the past years, then by the year of 2020, the width of a wire in a computer chip will be no more than a size of a single atom. These are sizes for which rules of classical physics no longer apply. Computers designed on today's chip technology will not continue to get cheaper and better. Because of its great power, quantum computer is an attractive next step in computer technology. (Manay, 1998, p. 5).

A technology of quantum computers is also very different. For operation, quantum computer uses quantum bits (*qubits)*. Qubit has a quaternary nature. Quantum mechanic's laws are completely different from the laws of a classical physics. A qubit can exist not only in the states corresponding to the logical values 0 or 1 as in the case of a classical bit, but also in a superposition state.

A qubit is a bit of information that can be both zero and one simultaneously (Superposition state). Thus, a computer working on a qubit rather than a standard bit can make calculations using both values simultaneously. A qubyte, is made up of eight qubits and can have all values from zero to 255 simultaneously. "Multi-qubyte systems have a power beyond anything possible with classical computers." (Quantum Computers & Moore's Law, p.1)

Forty qubits could have the same power as modern supercomputers. According to Chuang a supercomputer needs about a month to find a phone number from the database consisting of world's phone books, where a quantum computer is able to solve this task in 27 minutes.

Massachusetts Institute of Technology, Oxford University, IBM and Los Alamos National Laboratory are the most successful in development of quantum computer. (West, 2000, &7)

## History of Quantum Computers

In 1982 R.Feynman presented an interesting idea how the quantum system can be used for computation reasons. He also gave an explanation how effects of quantum physics could be simulated by such quantum computer. This was very interesting idea which can be used for future research of quantum effects. Every experiment investigating the effects and laws of quantum physics is complicated and expensive. Quantum computer would be a system performing such experiments permanently. Later in 1985, it was proved that a quantum computer would be much more powerful than a classical one. (West, 2000, p. 3)

## The Major Difference between Quantum and Classical Computers

The memory of a classical computer is a string of 0s and 1s, and it can perform calculations on only one set of numbers simultaneously. The memory of a quantum computer is a quantum state that can be a superposition of different numbers. A quantum computer can do an arbitrary reversible classical computation on all the numbers simultaneously. Performing a computation on many different numbers at the same time and then interfering all the results to get a single answer, makes a quantum computer much powerful than a classical one. (West, 2000)

## The Potential and Power of Quantum Computing

Quantum computer with 500 qubits gives $2^{500}$ superposition states. Each state would be classically equivalent to a single list of 500 1's and 0's. Such computer could operate on $2^{500}$ states simultaneously. Eventually, observing the system would cause it to collapse into a single quantum state corresponding to a single answer, a single list of 500 1's and 0's, as dictated by the measurement axiom of quantum mechanics. This kind of computer is equivalent to a classical computer with approximately $10^{150}$ processors. (West, 2000, p. 3)

## Moore's Law for Quantum Computers

According to Moore's Law, the number of transistors of a microprocessor continues to double in every 18 months. According to such evolution if there is a classical computer in year 2020, it will run at 40 GHz CPU speed with 160 Gb RAM. If we use an analogue of Moor's law for quantum computers, the number of quantum bits would be double in every 18 months. But adding just one qubit is already enough to double a speed. So, the speed of quantum computer will increase more than just doubling it. (Quantum Computers & Moore's Law, §1)

## Some Problems in Production of Quantum Computers

Any kind of measurement of quantum state parameters considers interaction process with environment (with other particles - particle of light for example), which causes a change of some parameters of this quantum state. Measurement of superposition quantum state will collapse it into a classical state. This is called decoherence. This is the major obstacle in a process of producing of a quantum computer. If decoherence problem cannot be solved, a quantum computer will be no better than a silicon one. (Daniel, 1999)

In order to make quantum computers powerful, many operations must be performed before quantum coherence is lost. It can be impossible to construct a quantum computer that will make calculations before decohering. But if one makes a quantum computer, where the number of errors is low enough, than it is possible to use an error-correcting code for preventing data looses even when qubits in the computer decohere. There are a lot of error-correcting codes. One of the simplest classical error-correcting codes is called repetition code. 0 is encoded as 000 and 1 as 111. Then if only one bit is flipped, one gets a state for example 011 that can be corrected to its original state 111. The signs of states in a quantum superposition are also important, but sign errors can also be corrected. There exists even a theory about quantum error-correcting codes. (Daniel, 1999, p. 1)

Another problem is hardware for quantum computers. Nuclear Magnetic Resonance (NMR) technology is the most popular today, because of some successful experiments. MIT and Los Alamos National Laboratory have constructed a simple quantum computer using NMR technology. Some other designs are based on ion trap and quantum electrodynamics (QED). All of these methods have significant limitations. Nobody knows what the architecture of future quantum computers hardware will be. (West, 2000, p. 6)

## Future Benefits of Quantum Computers

**1.** *Cryptography and Peter Shor's Algorithm*

In 1994 Peter Shor (Bell Laboratories) found out the first quantum algorithm that, in principle, can perform an efficient factorization. This became a complex application that only a quantum computer could do. Factoring is one of the most important problems in cryptography. For instance, the security of RSA (electronic banking security system) - public key cryptography - depends on factoring and it is a big problem. Because of many useful features of quantum computer, scientists put more efforts to build it. However, breaking any kind of current encryption that takes almost centuries on existing computers, may just take a few years on quantum computer. (Maney, 1998)

*2. Artificial Intelligence*

It has been mentioned that quantum computers will be much faster and consequently will perform a large amount of operations in a very short period of time. On the other side, increasing the speed of operation will help computers to learn faster even using the one of the simplest methods - mistake bound model for learning.

*3. Other Benefits*

High performance will allow us in development of complex compression algorithms, voice and image recognition, molecular simulations, true randomness and quantum communication. Randomness is important in simulations. Molecular simulations are important for developing simulation applications for chemistry and biology. With the help of quantum communication both receiver and sender are alerted when an eavesdropper tries to catch the signal. Quantum bits also allow more information to be communicated per bit. Quantum computers make communication more secure.

## Strange Thing about Quantum Computers

"On the theory side, quantum mechanics delves deep into areas that are nearly unthinkable. For instance, it's possible that a quantum computer holds an infinite number of right answers for an infinite number of parallel universes. It just happens to give you the right answer for the universe you happen to be in at the time. "It takes a great deal of courage to accept these things," says Charles Bennett of IBM, one of the best known quantum computing scientists. "If you do, you have to believe in a lot of other strange things." " (Manay, 1998)

## Dancing Chloroform Atoms

A few years ago, Gershenfeld and Chuang made the first quantum computer. It was based on nuclear magnetic resonance technology. The program was performing a simple search using Grover's algorithm. In comparison to classical computers it took one item out of four in just one step, instead of making two or three steps as classical computes. The price for making the first 2-qubit computer was approximately $1 million.

## Entanglement of Quantum Systems

According to quantum mechanics an outside force acting on two particles of the quantum system can cause them to become entangled. The quantum state of this system can contain all positions of spins (internal magnetic moments) of each particle. The total spin of the system can only be equal to certain discrete values with different probabilities. Measurements of total spin of certain quantum systems showed that positions of spins of some particle are not independent from others. For such systems, when an orientation of a spin of one particle changed by some reason, an orientation of a spin of another particle changes automatically and instantly. The laws that that have been developed so far about the speed of light are disobeyed in this case, because the change in an orientation of a spin happens immediately. At least there is hypothesize to use this phenomena for quantum computing.

It is well known that a speed of communication is limited by a speed of light as nothing can travel faster than the speed of light. The question is how particles of the quantum system communicate when they change their spin orientation and consequently their vector states. Famous scientists spent a lot of time discussing this issue. Einstein's idea that some unknown "hidden parameters" of quantum system were contributing to this effect has been rejected theoretically and experimentally.

This is one of the example showing the difference between classical and quantum realities. This effect of the quantum system explains a lot of aspects of the nature (f.e. chemical characteristics of atoms and molecules) and is proved by the experiments.

"In fact, theories about entanglement have led scientists to believe there could be a way to speed up computing. Even today's computers are nearing a point at which their speed is being limited by how fast an electron can move through a wire - the speed of light. Whether in a quantum or traditional computer, entanglement could blow past that limit." (Manay, 1998)

## Conclusion

It is important that making a practical quantum computing is still far in the future. Programming style for a quantum computer will also be quite different. Development of quantum computer needs a lot of money. Even the best scientists can't answer a lot of questions about quantum physics. Quantum computer is based on theoretical physics and some experiments are already made. Building a practical quantum computer is just a matter of time. Quantum computers easily solve applications that can't be done with help of today's computers. This will be one of the biggest steps in science and will undoubtedly revolutionize the practical computing world.